\documentclass{article} 
\usepackage{iclr2027_conference,times}

\usepackage{amsmath,amsfonts,bm}

\def\eqref#1{equation~\ref{#1}}

\def\1{\bm{1}}

\DeclareMathAlphabet{\mathsfit}{\encodingdefault}{\sfdefault}{m}{sl}
\SetMathAlphabet{\mathsfit}{bold}{\encodingdefault}{\sfdefault}{bx}{n}

\usepackage{hyperref}
\usepackage{url}
\usepackage{booktabs}
\usepackage{graphicx}
\usepackage{multirow}
\usepackage[table]{xcolor}
\usepackage{wrapfig}
\usepackage{algorithm}
\usepackage{algorithmic}
\usepackage{amssymb} 

\title{Attn-Forcing: Learning Full-Context Attention Proxies for Streaming Video Generation}
\iclrfinalcopy

\author{{Junyao Gao}$^{1}$\thanks{Equal contributions. \textsuperscript{\ddag} corresponding authors. Work done during Junyao Gao's internship at Tencent.}, {Sibo Liu}$^{2}$\textsuperscript{*}, {Jiaxing Li}$^{3}$\textsuperscript{*}, {Yanan Sun}$^{4}$, {Hanxin Zhu}$^{2}$\\
\textbf{{Weidong Zhang}}$^{2}$\textbf{,} \textbf{{Jun Zhang}}$^{2}$\textsuperscript{\ddag}\textbf{,} \textbf{{Cairong Zhao}}$^{1}$\footnotemark[2]\textsuperscript{\ddag}\\
$^{1}$Tongji University, $^{2}$Tencent, $^{3}$Nanyang Technological University, \\ $^{4}$Hong Kong University of Science and Technology\\
}
\begin{document}

\maketitle
\vspace{-6mm}
\begin{abstract}
\vspace{-2mm}
Increasing the attention context length significantly improves streaming video generation quality, while incurring prohibitive memory and computational costs as the context grows.
Existing methods attempt to capture this contextual information through attention selection or compression, but remain limited by bounded representations and lack explicit attention-level supervision to retain historical contributions to current queries.
To address this, we introduce \textbf{Attn-Forcing}, a novel framework that preserves full-context attention contributions for streaming video generation under bounded memory and computation.
We achieve this by introducing attention proxies that continuously approximate the attention responses induced by the historical context.
Specifically, we implement the proxies as learnable visual latents and employ a Proxy Update Module to incorporate visual context from historical latents.
Moreover, we propose an \textbf{Attention Matching Loss} to explicitly constrain the attention behavior of the learned proxies using full-context attention supervision.
Extensive qualitative and quantitative experiments demonstrate that Attn-Forcing consistently improves streaming video generation quality and achieves state-of-the-art performance over existing baseline methods.
\end{abstract}

\section{Introduction}
Streaming video generation has gained increasing attention due to its broad applications in interactive content creation, digital humans, live streaming, and persistent virtual environments.
With recent advances in autoregressive (AR) diffusion models \citep{jin2025pyramidal, chen2025skyreels, wu2025pack, yuan2026helios, teng2025magi, ma2026flex}, streaming generation is commonly implemented through chunk-by-chunk generation with causal KV caching in Diffusion Transformers (DiT) \citep{peebles2023scalable}.

\begin{wrapfigure}{r}{0.6\linewidth}
		\centering
		\includegraphics[width=\linewidth]{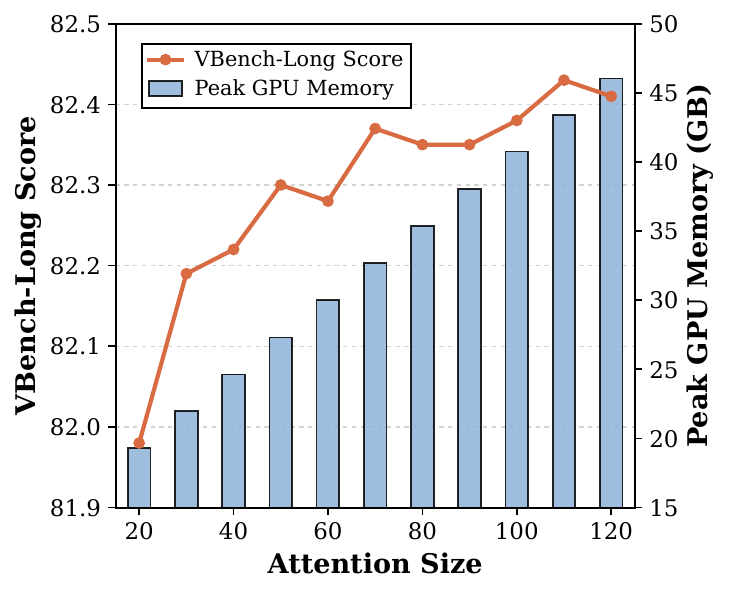}
        \vspace{-8mm}
\caption{VBench-Long score and peak GPU memory of Self-Forcing \citep{huang2026self} during 30s video inference with different local attention sizes. A larger attention window generally improves generation quality, but at the cost of increased peak GPU memory.}
\label{fig:intro}
\end{wrapfigure}

Existing methods \citep{huang2026self, zhu2026causal, yi2025deep, liu2026rolling, yang2025longlive} typically retain only a limited-length KV cache during training and inference.
This truncation of full-context attention discards contributions from historical KV states, leading to degraded generation quality.
As shown by the experiment in Figure \ref{fig:intro}, although Self-Forcing \citep{huang2026self} is trained with a fixed local attention window, simply increasing the KV cache length at inference consistently yields higher VBench-Long \citep{huang2024vbench} scores, highlighting the benefit of broader historical context.
However, under bounded memory and computational budgets, maintaining increasingly long KV contexts is impractical for streaming generation.

To address this challenge, \cite{guo2025end,yi2025deep,ji2025memflow,ma2026flow} selectively retain or retrieve relevant historical KV states through adaptive strategies, but remain limited by bounded KV budgets and potentially accumulated errors in current visual latents.
Another line of work \citep{wu2025pack,lu2026reward,kim2026memrope,wu2026infinite,xue2026ring, bian2026echo} compresses historical context into compact representations using dedicated objectives or predefined compression ratios/schedules, but inevitably faces a trade-off between context coverage and information fidelity.
Fundamentally, these methods still perform attention over bounded historical representations and lack attention-level supervision, making it difficult to preserve the contribution of the full context to current queries.

Can we capture this contribution without explicitly maintaining the full context?

In this paper, we answer this question by proposing \textbf{Attn-Forcing}, a novel framework that approximates full-context attention under bounded memory and computation.
We achieve this by learning additional attention proxies during distribution matching distillation (DMD) training \citep{yin2024one}.
Specifically, we implement the proxies as learnable latents and continuously update them through a \textbf{Proxy Update Module} using visual latents evicted from the local attention window to incorporate the visual history.
We then forward the proxy latents through the DiT and use their layer-wise KV states for subsequent streaming generation.
However, the DMD training objective primarily matches the model output distribution, without explicitly constraining the attention behavior induced by the learned proxies.
We therefore introduce an \textbf{Attention Matching Loss} to directly match the attention responses obtained with the learned proxies to those of the historical attention.

Extensive qualitative and quantitative experiments demonstrate that Attn-Forcing achieves state-of-the-art (SOTA) performance over baseline methods.
Moreover, ablation studies  confirm the effectiveness and advantages of proposed designs, providing valuable insights for the streaming video generation community.

The contributions of this paper are summarized as follows:
\begin{itemize}
    \item We propose Attn-Forcing, a novel DMD training framework that, for the first time, preserves full-context attention contributions under bounded memory and computation during streaming generation.
    \item We introduce learnable attention proxies to continuously incorporate visual history and a novel Attention Matching Loss to further align their attention responses with full-context attention.
    \item Experiments show that Attn-Forcing improves streaming video generation quality and achieves SOTA performance over baseline methods.
\end{itemize}

\section{Related Work}
\subsection{Video Diffusion Models}
Early video diffusion models typically extend image diffusion backbones with temporal modules to model spatiotemporal dependencies, enabling high-quality text-to-video generation \citep{ho2022video,blattmann2023stable,chen2023videocrafter1,guo2023animatediff}.
With the emergence of Diffusion Transformers (DiT) \citep{peebles2023scalable}, video generation has progressively shifted toward large-scale transformer architectures with stronger scalability and global temporal modeling.
Representative models such as Sora \citep{videoworldsimulators2024}, CogVideoX \citep{yang2025cogvideox}, MovieGen \citep{polyak2024movie}, HunyuanVideo \citep{kong2024hunyuanvideo}, and Wan \citep{wan2025wan} have demonstrated impressive video generation capabilities, producing high-fidelity videos with coherent motion, strong temporal consistency, and faithful alignment with textual prompts.
Recent open-source models, including LTX-Video \citep{hacohen2024ltx}, Wan2.2 \citep{wan2025wan}, and HunyuanVideo-1.5 \citep{wu2025hunyuanvideo}, further push the performance and efficiency of open-source video generation.
Meanwhile, frontier systems such as Veo 3.1 \citep{google2025veo31}, Kling 3.0 \citep{kuaishou2026kling3}, Seedance 2.0 \citep{seedance2026seedance}, MiniMax H3 \citep{minimax2026h3}, and HappyHorse \citep{alibaba2026happyhorse} exhibit increasingly sophisticated generation capabilities, including realistic physical dynamics, complex motion, multimodal conditioning, and joint audio-video generation.
Our work is built upon Wan2.1-T2V-1.3B \citep{wan2025wan}, adapting it for streaming long video generation.

\subsection{Long Video Generation}
Long video generation remains challenging for video diffusion models due to the high computational cost of full-sequence modeling and the temporal extrapolation gap between short-video training and long-video inference.
One line of work seeks to extrapolate pretrained short-video diffusion models to longer durations without costly long-video retraining.
Representative approaches modify the inference process or temporal modeling through noise rescheduling, progressive denoising, or positional extrapolation, including FreeNoise \citep{qiu2024freenoise}, FIFO-Diffusion \citep{kim2024fifo}, RIFLEx \citep{zhao2025riflex}, and Diff-VF \citep{yang2026diff}.
These methods substantially extend the generation horizon of pretrained video diffusion models while improving temporal continuity over long sequences.

More recently, autoregressive (AR) video diffusion has emerged as a promising paradigm for streaming long video generation.
Specifically, CausVid \citep{yin2025slow} introduces causal diffusion distillation to enable efficient streaming generation, while Self-Forcing \citep{huang2026self} further trains on self-generated histories to reduce the autoregressive train-test discrepancy.
Subsequent works such as LongLive \citep{yang2025longlive} and Self-Forcing++ \citep{cui2026self} further improve long-horizon generation through dedicated long-rollout training and inference strategies.
However, due to GPU memory constraints, these methods are restricted to fixed-length KV caches that continuously discard historical information, resulting in degraded and unstable generation quality.

To address this, recent works explore different strategies for preserving useful historical context under bounded memory and computation.
Resampling Forcing \citep{guo2025end} and MemFlow \citep{ji2025memflow} dynamically retrieve or activate relevant historical states for current generation, while Deep Forcing \citep{yi2025deep} and FlowCache \citep{ma2026flow} prune or compress KV caches according to their estimated importance.
Other approaches \citep{wu2025pack,lu2026reward,kim2026memrope,xue2026ring,bian2026echo} aggregate or compress historical information into compact representations with bounded computational cost.
However, these methods still perform attention over bounded historical representations and lack explicit attention-level supervision to preserve full-context contributions to current queries.
To address this, we introduce attention proxies and train them with an additional Attention Matching loss.

\section{Method}
\begin{figure*}
    \centering
    \includegraphics[width=1.0\linewidth]{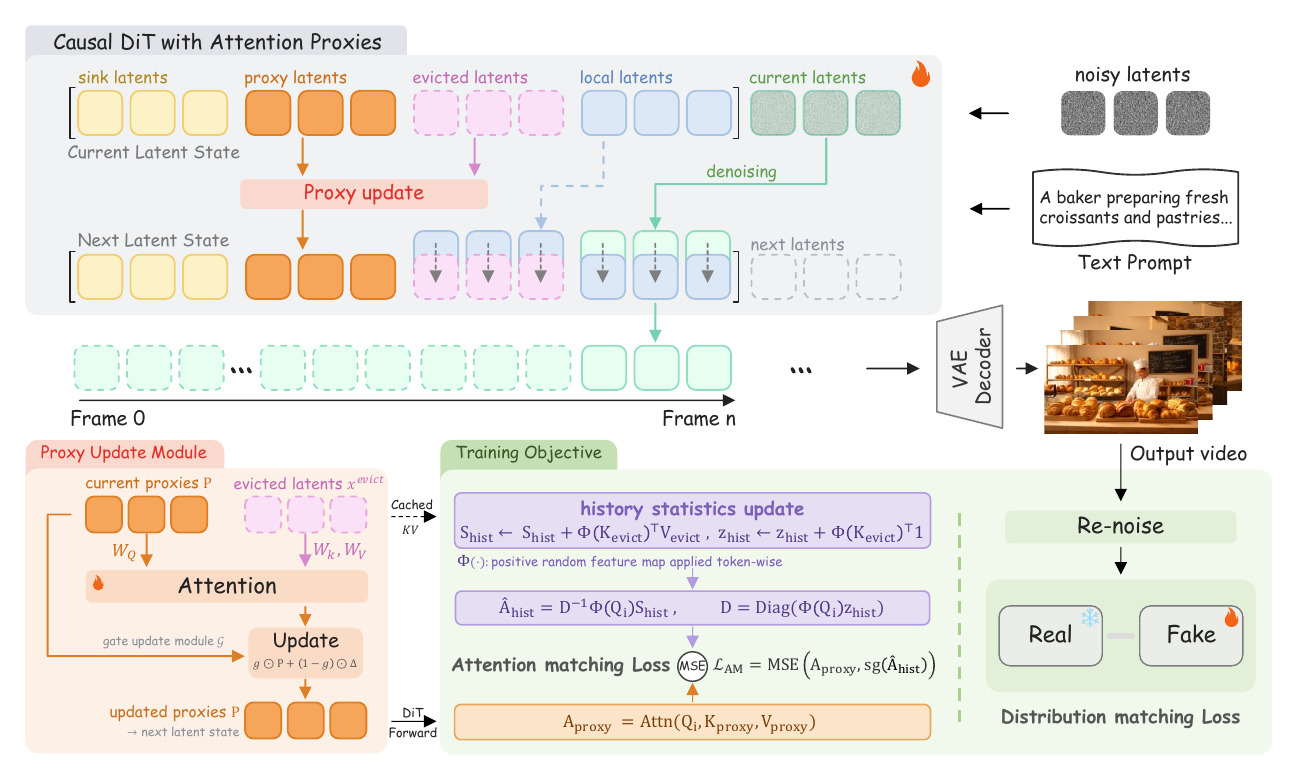}
    \vspace{-3mm}
    \caption{The framework of Attn-Forcing. During training, learnable proxy latents are continuously updated using visual latents evicted from the local attention window through a Proxy Update Module. Attention Matching Loss further aligns the proxy-induced attention responses with historical attention responses approximated using online accumulated statistics.}
     \vspace{-4mm}
    \label{fig:framework}
\end{figure*}

In this section, we present the technical details of the proposed Attn-Forcing.
The overall framework is illustrated in Figure~\ref{fig:framework}.
We first introduce the preliminaries of KV caching and distribution matching distillation in Section~\ref{sec:pre}.
We then present the construction of attention proxies and the Attention Matching Loss in Sections~\ref{sec:pro} and~\ref{sec:aml}, respectively.

\subsection{Preliminary}
\label{sec:pre}
\noindent{\textbf{KV Cache.}}
Streaming video generation with autoregressive Diffusion Transformers predicts each new video chunk through causal attention over previous chunks, whose key and value states are cached and reused for current generation:
\begin{equation}
\operatorname{Attn}
\left(
\mathbf Q_i,
[\mathbf K_{\mathrm{sink}},\mathbf K_{\mathrm{local}}],
[\mathbf V_{\mathrm{sink}},\mathbf V_{\mathrm{local}}]
\right)
=
\operatorname{Softmax}
\left(
\frac{
\mathbf Q_i
[\mathbf K_{\mathrm{sink}},\mathbf K_{\mathrm{local}}]^\top
}{
\sqrt{d_h}
}
\right)
[\mathbf V_{\mathrm{sink}},\mathbf V_{\mathrm{local}}],
\label{eq:attn}
\end{equation}
where $\mathbf Q_i$ denotes the queries of the $i$-th video chunk, while $(\mathbf K_{\mathrm{sink}},\mathbf V_{\mathrm{sink}})$ and $(\mathbf K_{\mathrm{local}},\mathbf V_{\mathrm{local}})$ denote the cached sink and local key-value states, respectively.

\noindent{\textbf{Distribution Matching Distillation (DMD).}}
Following prior methods \citep{yin2024one}, we adopt the DMD objective to distill a multi-step bidirectional diffusion model into a few-step causal generator. Specifically, given the $i$-th generated video chunk $\mathbf{x}_i$, we perturb it at diffusion timestep $t$ as
\begin{equation}
\mathbf{x}_{i,t}
=
\alpha_t \mathbf{x}_i
+
\sigma_t \boldsymbol{\epsilon},
\qquad
\boldsymbol{\epsilon}\sim\mathcal{N}(\mathbf{0},\mathbf{I}),
\end{equation}
and estimate its score using the pretrained diffusion teacher and a learnable fake-score model. The DMD gradient for the causal generator is given by
\begin{equation}
\label{eq:dmd}
\nabla_{\theta}\mathcal{L}_{\mathrm{DMD}}
=
\mathbb{E}_{i,t,\boldsymbol{\epsilon}}
\left[
w_t
\left(
\mathbf{s}_{\mathrm{fake}}(\mathbf{x}_{i,t},t)
-
\mathbf{s}_{\mathrm{real}}(\mathbf{x}_{i,t},t)
\right)
\frac{\partial \mathbf{x}_i}{\partial \theta}
\right],
\end{equation}
where $\mathbf{s}_{\mathrm{real}}$ and $\mathbf{s}_{\mathrm{fake}}$ denote the score estimates from the pretrained diffusion teacher and the fake-score model, respectively, and $w_t$ denotes the timestep-dependent weighting factor.

\subsection{Attention Proxies}
\label{sec:pro}
As discussed in the Introduction and Figure~\ref{fig:intro}, retaining a longer KV context improves generation quality.
However, due to memory and computational constraints, prior methods \citep{huang2026self, yang2025longlive} restrict the KV cache in Eq.~\ref{eq:attn} to a fixed-length local window during training and inference.
In this setting, older historical states are continuously evicted as generation proceeds, thereby discarding their attention contributions to current queries.
Accordingly, to preserve such historical attention contributions without explicitly retaining the growing historical KV cache, we introduce proxy latents to approximate their attention responses to the current query $\mathbf{Q}_i$:
\begin{equation}
\operatorname{Attn}
\left(
\mathbf{Q}_i,
\mathbf{K}_{\mathrm{proxy}},
\mathbf{V}_{\mathrm{proxy}}
\right)
\approx
\operatorname{Attn}
\left(
\mathbf{Q}_i,
\mathbf{K}_{\mathrm{hist}},
\mathbf{V}_{\mathrm{hist}}
\right),
\label{eq:tgt}
\end{equation}
where $(\mathbf{K}_{\mathrm{hist}}, \mathbf{V}_{\mathrm{hist}})$ denote the evicted historical KV states and $(\mathbf{K}_{\mathrm{proxy}}, \mathbf{V}_{\mathrm{proxy}})$ denote the key and value states derived from the proxy latents.

Specifically, as shown in Figure \ref{fig:framework}, we implement these proxies as $K=3$ learnable latents in the same feature space as the DiT visual latents.
They are randomly initialized and continuously updated using visual latents evicted from the local attention window as follows.
\begin{equation}
\mathbf{P}
\leftarrow
\mathcal{U}
\left(
\mathbf{P},
\mathbf{x}^{\mathrm{evict}}
\right)
=
\mathcal{G}\left(
\mathbf{P},
\operatorname{Attn}
\left(
\mathbf{W}_{Q}(\mathbf{P}),
\mathbf{W}_{K}(\mathbf{x}^{\mathrm{evict}}),
\mathbf{W}_{V}(\mathbf{x}^{\mathrm{evict}})
\right)
\right),
\label{eq:proxy_update}
\end{equation}
where $\mathcal{U}$ denotes the Proxy Update Module, and $\mathbf{W}_{Q}(\cdot)$, $\mathbf{W}_{K}(\cdot)$, and $\mathbf{W}_{V}(\cdot)$ denote its learnable query, key, and value projections, respectively.
For the newly evicted visual latents $\mathbf{x}^{\mathrm{evict}}$, the current proxy latents $\mathbf{P}$ query the evicted latents through cross-attention to capture their contextual information, which is then further processed and fused with $\mathbf{P}$ through the gated update module $\mathcal{G}$. More details are provided in Appendix \ref{app:proxy_update}.
The updated proxy latents are then propagated through the DiT to obtain the layer-wise key and value states for subsequent chunk generation:
\begin{equation}
\mathbf{K}^{l}_{\mathrm{proxy}}
=
\mathbf{W}_{K}^{l}(\mathbf{P}^{l}),
\qquad
\mathbf{V}^{l}_{\mathrm{proxy}}
=
\mathbf{W}_{V}^{l}(\mathbf{P}^{l}),
\label{eq:proxy_kv}
\end{equation}
where $\mathbf{P}^{0}=\mathbf{P}$ and $\mathbf{P}^{l+1}=B^{l}(\mathbf{P}^{l})$, with $B^{l}$ denoting the $l$-th DiT block, and $\mathbf{W}_{K}^{l}(\cdot)$ and $\mathbf{W}_{V}^{l}(\cdot)$ denoting its key and value projections.
The resulting proxy KV states are then denoted as $\mathbf{K}_{\mathrm{proxy}}, \mathbf{V}_{\mathrm{proxy}}$ and appended between the sink and local KV states in Eq.~\ref{eq:attn}, yielding
\begin{equation}
\operatorname{Attn}
\left(
\mathbf Q_i,
[\mathbf K_{\mathrm{sink}},\mathbf K_{\mathrm{proxy}},\mathbf K_{\mathrm{local}}],
[\mathbf V_{\mathrm{sink}},\mathbf V_{\mathrm{proxy}},\mathbf V_{\mathrm{local}}]
\right).
\label{eq:proxy_attn}
\end{equation}
The proxy KV states allow evicted visual context to continue contributing to the attention responses of current queries without explicitly retaining the full historical KV cache.


\subsection{Attention Matching Loss}
\label{sec:aml}
Although the proxy latents continuously incorporate visual context from the evicted latents, their induced attention responses may not faithfully  reproduce those induced by the full historical KV states.
This is because the standard DMD objective used during training (Eq.~\ref{eq:dmd}) mainly focuses on distribution-level score matching, rather than directly aligning the attention responses in Eq.~\ref{eq:tgt}.
However, computing historical attention responses requires explicitly retaining these historical states, resulting in unacceptable memory and computational costs as the historical context grows.
To address this, we propose an efficient approximation to
$\operatorname{Attn}\left(\mathbf{Q}_i,\mathbf{K}_{\mathrm{hist}},\mathbf{V}_{\mathrm{hist}}\right)$
using online accumulated statistics, avoiding growing memory and computational costs.

Specifically, we first rewrite the attention response for each query
$\mathbf{q}\in\mathbf{Q}_i$ in
$\operatorname{Attn}\left(\mathbf{Q}_i,\mathbf{K}_{\mathrm{hist}},\mathbf{V}_{\mathrm{hist}}\right)$
as
\begin{equation}
\mathbf{a}_{\mathrm{hist}}(\mathbf{q})
=
\frac{
\sum_{j\in\mathcal{H}}
\kappa(\mathbf{q},\mathbf{k}_{j})\mathbf{v}_{j}
}{
\sum_{j\in\mathcal{H}}
\kappa(\mathbf{q},\mathbf{k}_{j})
},
\qquad
\kappa(\mathbf{q},\mathbf{k})
=
\exp\left(
\frac{\mathbf{q}^{\top}\mathbf{k}}{\sqrt{d_h}}
\right),
\label{eq:hist_attn_exact}
\end{equation}
where $\mathcal{H}$ indexes the KV states in
$(\mathbf{K}_{\mathrm{hist}},\mathbf{V}_{\mathrm{hist}})$.
Next, following \cite{choromanski2020rethinking}, we leverage the positive random feature representation of the softmax kernel.
Given a positive random feature $\phi_{\omega}(\cdot)$, the softmax kernel can be expressed as
\begin{equation}
\kappa(\mathbf{q},\mathbf{k})
=
\mathbb{E}_{\omega}
\left[
\phi_{\omega}(\mathbf{q})
\phi_{\omega}(\mathbf{k})
\right].
\label{eq:kernel_expectation}
\end{equation}
Approximating this expectation with $M$ sampled random features gives
\begin{equation}
\kappa(\mathbf{q},\mathbf{k})
\approx
\phi(\mathbf{q})^{\top}\phi(\mathbf{k}),
\label{eq:kernel_approx}
\end{equation}
where $\phi(\cdot)\in\mathbb{R}^{M}$ denotes the resulting finite-dimensional positive random feature mapping, with $M$ being the number of random features and the attention scaling absorbed into $\phi(\cdot)$.
A detailed derivation is provided in Appendix~\ref{app:prf}.
Substituting Eq.~\ref{eq:kernel_approx} into Eq.~\ref{eq:hist_attn_exact} gives
\begin{equation}
\mathbf{a}_{\mathrm{hist}}(\mathbf{q})
\approx
\frac{
\sum_{j\in\mathcal{H}}
\phi(\mathbf{q})^{\top}\phi(\mathbf{k}_{j})\mathbf{v}_{j}
}{
\sum_{j\in\mathcal{H}}
\phi(\mathbf{q})^{\top}\phi(\mathbf{k}_{j})
}
=
\frac{
\left(
\sum_{j\in\mathcal{H}}
\phi(\mathbf{k}_{j})\mathbf{v}_{j}^{\top}
\right)^{\top}
\phi(\mathbf{q})
}{
\left(
\sum_{j\in\mathcal{H}}
\phi(\mathbf{k}_{j})
\right)^{\top}
\phi(\mathbf{q})
}.
\label{eq:hist_attn_factorized}
\end{equation}
This factorization allows the historical attention response to be approximated using two online accumulated statistics:
\begin{equation}
\mathbf{S}_{\mathrm{hist}}
=
\sum_{j\in\mathcal{H}}
\phi(\mathbf{k}_{j})\mathbf{v}_{j}^{\top},
\qquad
\mathbf{z}_{\mathrm{hist}}
=
\sum_{j\in\mathcal{H}}
\phi(\mathbf{k}_{j}).
\label{eq:hist_statistics}
\end{equation}
Whenever new KV states are evicted from the local attention window, these statistics are incrementally updated as
\begin{align}
\mathbf{S}_{\mathrm{hist}}
&\leftarrow
\mathbf{S}_{\mathrm{hist}}
+
\Phi(\mathbf{K}_{\mathrm{evict}})^{\top}
\mathbf{V}_{\mathrm{evict}},
\nonumber\\
\mathbf{z}_{\mathrm{hist}}
&\leftarrow
\mathbf{z}_{\mathrm{hist}}
+
\Phi(\mathbf{K}_{\mathrm{evict}})^{\top}\mathbf{1},
\label{eq:hist_statistics_update}
\end{align}
where $\Phi(\cdot)$ applies $\phi(\cdot)$ token-wise.
The sizes of $\mathbf{S}_{\mathrm{hist}}$ and $\mathbf{z}_{\mathrm{hist}}$ are independent of the historical context length, preventing the memory and attention computation from growing with the accumulated history.

Extending Eq.~\ref{eq:hist_attn_factorized} to all queries in $\mathbf{Q}_i$, we obtain
\begin{equation}
\widehat{\mathbf{A}}_{\mathrm{hist}}
=
\mathbf{D}^{-1}
\Phi(\mathbf{Q}_i)
\mathbf{S}_{\mathrm{hist}},
\qquad
\mathbf{D}
=
\operatorname{Diag}
\left(
\Phi(\mathbf{Q}_i)\mathbf{z}_{\mathrm{hist}}
\right),
\label{eq:hist_attn_approx}
\end{equation}
where
$\widehat{\mathbf{A}}_{\mathrm{hist}}$
provides an efficient approximation to
$\operatorname{Attn}\left(
\mathbf{Q}_i,
\mathbf{K}_{\mathrm{hist}},
\mathbf{V}_{\mathrm{hist}}
\right)$
in Eq.~\ref{eq:tgt}.
Since $\widehat{\mathbf{A}}_{\mathrm{hist}}$ is obtained via random-feature approximation rather than native KV states, directly using it for generation would introduce approximation errors into the attention computation.
We therefore use it only as an attention-level supervision target.

For the proxy KV states, the corresponding attention response is
\begin{equation}
\mathbf{A}_{\mathrm{proxy}}
=
\operatorname{Attn}
\left(
\mathbf{Q}_i,
\mathbf{K}_{\mathrm{proxy}},
\mathbf{V}_{\mathrm{proxy}}
\right).
\label{eq:proxy_attn_response}
\end{equation}
We then define the Attention Matching Loss as
\begin{equation}
\mathcal{L}_{\mathrm{AM}}
=
\operatorname{MSE}
\left(
\mathbf{A}_{\mathrm{proxy}},
\operatorname{sg}
\left(
\widehat{\mathbf{A}}_{\mathrm{hist}}
\right)
\right).
\label{eq:attention_matching_loss}
\end{equation}
where $\operatorname{sg}(\cdot)$ denotes stop-gradient.
Since $\widehat{\mathbf{A}}_{\mathrm{hist}}$ approximates the attention response induced by the full historical KV states, minimizing $\mathcal{L}_{\mathrm{AM}}$ provides a tractable surrogate for directly optimizing the approximation objective in Eq.~\ref{eq:tgt}.
In practice, we apply this objective to the last $L$ DiT layers and average the resulting losses across them.

Together with the DMD objective, the overall training objective is defined as:
\begin{equation}
\mathcal{L}
=
\mathcal{L}_{\mathrm{DMD}}
+
\lambda_{\mathrm{AM}}\mathcal{L}_{\mathrm{AM}},
\label{eq:overall_loss}
\end{equation}
where $\lambda_{\mathrm{AM}}$ controls the strength of attention matching.
The DMD objective jointly optimizes the generator $G_\theta$ and proxy-related parameters, while the Attention Matching Loss only optimizes the proxy-related parameters.
More details are provided in Appendix~\ref{app:training_inference}.

\begin{table}[t]
    \centering
    \setlength{\tabcolsep}{5pt}
    \renewcommand{\arraystretch}{1.0}
    \caption{Quantitative comparison on 5s short videos and 30s long videos generated from the official VBench prompts. Best results are shown in \textbf{bold}, and second-best results are \underline{underlined}.}

    \resizebox{\linewidth}{!}{
    \begin{tabular}{@{}lcc ccc ccc@{}}
        \toprule
        \multirow{2}{*}{Model}
        & \multirow{2}{*}{\#Params}
        & \multirow{2}{*}{Resolution}
        & \multicolumn{3}{c}{Evaluation Scores on \textbf{5s} $\uparrow$}
        & \multicolumn{3}{c}{Evaluation Scores on \textbf{30s} $\uparrow$} \\
        \cmidrule(lr){4-6}
        \cmidrule(l){7-9}
        & & 
        & Total & Quality & Semantic
        & Total & Quality & Semantic \\
        \midrule

        \rowcolor{gray!20}
        \multicolumn{9}{@{}l}{\textit{Diffusion models}} \\

        LTX-Video
        & 1.9B & $768{\times}512$
        & 79.31 & 81.51 & 70.53
        & -- & -- & -- \\

        Wan-2.1
        & 1.3B & $832{\times}480$
        & 82.84 & 83.64 & 79.67
        & -- & -- & -- \\

        \midrule
        \rowcolor{gray!20}
        \multicolumn{9}{@{}l}{\textit{Autoregressive models}} \\

        Self-Forcing
        & 1.3B & $832{\times}480$
        & \underline{83.76} & 84.43 & \textbf{81.06}
        & 81.92 & 82.98 & 77.68 \\

        Causal-Forcing
        & 1.3B & $832{\times}480$
        & 83.75 & \underline{84.58} & 80.41
        & 80.56 & 83.23 & 69.90 \\

        LongLive
        & 1.3B & $832{\times}480$
        & 83.53 & 84.23 & 80.71
        & \underline{83.15} & \underline{84.18} & \textbf{79.05} \\

        \midrule
        Attn-Forcing
        & 1.3B & $832{\times}480$
        & \textbf{84.10} & \textbf{84.94} & \underline{80.74}
        & \textbf{83.85} & \textbf{85.26} & \underline{78.18} \\

        \bottomrule
    \end{tabular}
    }
    \label{tab:vbench}
\end{table}

\section{Experiments}

\subsection{Implementation Details}
We build Attn-Forcing upon Wan2.1-T2V-1.3B \citep{wan2025wan}, which generates 5s clips at $832\times480$ resolution.
We first adopt the causal init model from Self-Forcing \citep{huang2026self} and apply DMD training on it to enable the frame sink and attention proxies.
We then perform streaming long tuning on 60-second sequences following LongLive \citep{yang2025longlive}, using the overall training objective in Eq.~\ref{eq:overall_loss}.
Text prompts are sampled from the filtered and LLM-augmented VidProM \citep{wang2024vidprom} dataset.
During streaming long tuning, we use $K=3$ proxy latents and $M=128$ random features, and apply the Attention Matching Loss to the last 6 DiT layers with $\lambda_{\mathrm{AM}}=5\times10^{-4}$.
We use four denoising steps $[1000,750,500,250]$ for each generated chunk, with a frame sink of 3 frames and an effective local attention window of 9 frames.
We adopt AdamW with learning rates of $1\times10^{-5}$, $2\times10^{-6}$, and $1\times10^{-4}$ for the generator, fake-score model, and attention proxies, respectively, and update the generator every 5 fake-score updates.
Training uses a total batch size of 32 on 16 H20 GPUs.
We train the short initialization stage for 700 steps and the long streaming stage for 2,300 steps, for a total of 3,000 training steps.

\begin{figure*}
    \centering
    \includegraphics[width=1.0\linewidth]{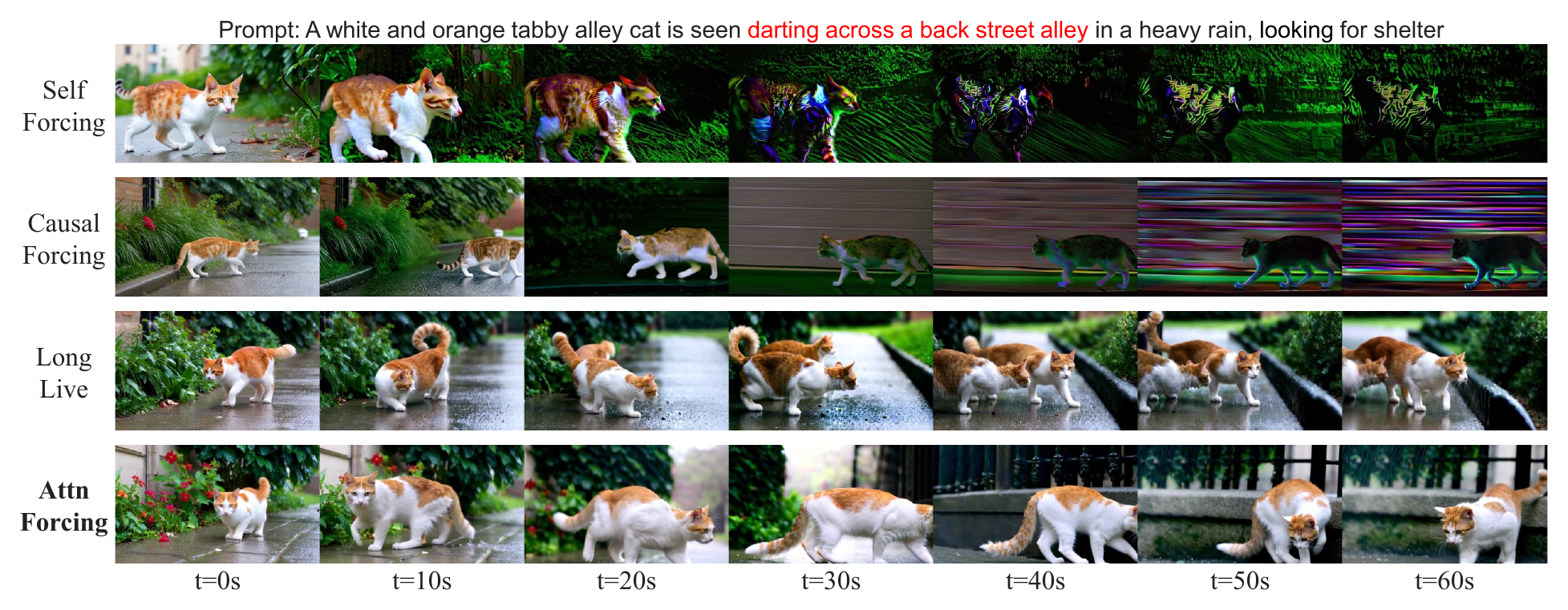}
    \vspace{-8mm}
    \caption{Qualitative comparison on 60s videos generated using MovieGen prompts. Attn-Forcing achieves stronger generation quality and motion dynamics. }
     \vspace{-4mm}
    \label{fig:main_result}
\end{figure*}

\noindent{\textbf{Evaluation Benchmarks \& Metrics.}}
Following the standard VBench evaluation protocol \citep{huang2024vbench}, we measure generation quality on both 5s and 30s videos using 946 prompts (each sampled with 5 different seeds) across 16 dimensions, with VBench used for 5s videos and VBench-Long for 30s videos.
Moreover, for longer-horizon evaluation, we follow \cite{yin2025slow} and generate 60s videos using 128 prompts from MovieGen \citep{polyak2024movie}, evaluating them using VBench-Long across six dimensions.

\begin{table}[t]
    \centering
    \setlength{\tabcolsep}{5pt}
    \renewcommand{\arraystretch}{1.0}
    \caption{Quantitative comparison on 60s long videos generated from the official MovieGen prompts. Best results are shown in \textbf{bold}, and second-best results are \underline{underlined}.}
    \label{tab:vbench_long}

    \resizebox{\linewidth}{!}{
    \begin{tabular}{@{}lccccccc@{}}
        \toprule
        \multirow{2}{*}{Model}
        & \multicolumn{7}{c}{Evaluation Scores on \textbf{60s} $\uparrow$} \\
        \cmidrule(l){2-8}
        & Total
        & Subject
        & Background
        & Smoothness
        & Dynamic
        & Aesthetic
        & Imaging Quality \\
        \midrule

        Self-Forcing
        & 78.29
        & 97.07
        & \underline{96.17}
        & 98.51
        & 34.53
        & 54.28
        & 67.32 \\

        Causal-Forcing
        & 79.05
        & 95.78
        & 95.80
        & 97.36
        & \underline{61.20}
        & 50.46
        & 64.78 \\

        LongLive
        & \underline{79.26}
        & \textbf{97.93}
        & \textbf{96.60}
        & \textbf{98.85}
        & 35.55
        & \underline{55.99}
        & \underline{68.81} \\

        \midrule
        Attn-Forcing
        & \textbf{82.37}
        & \underline{97.11}
        & 95.84
        & \underline{98.56}
        & \textbf{72.14}
        & \textbf{56.03}
        & \textbf{69.42} \\

        \bottomrule
    \end{tabular}
    }
\end{table}

\subsection{Comparison with SOTA Methods}
\noindent{\textbf{Short-Horizon (5s).}}
We compare Attn-Forcing against relevant open-source video generation models of similar scale, including diffusion-based models LTX-Video \citep{hacohen2024ltx} and Wan2.1 \citep{wan2025wan}, and autoregressive models Self-Forcing \citep{huang2026self}, Causal-Forcing \citep{zhu2026causal}, and LongLive \citep{yang2025longlive}.
We first report the quality, semantic, and total VBench scores on 5s videos in Table \ref{tab:vbench}. 
Attn-Forcing achieves the best quality and total scores at 84.94 and 84.10, respectively, and the second-best semantic score at 80.74, demonstrating superior overall video generation quality.

\noindent{\textbf{Long-Horizon (30s/60s).}}
For long-horizon evaluation, we generate 30s and 60s videos for all compared autoregressive methods using the VBench and MovieGen prompts, respectively.
We evaluate the 30s videos across all 16 VBench-Long dimensions, and the 60s videos across six VBench-Long dimensions: subject consistency, background consistency, motion smoothness, dynamic degree, aesthetic quality, and imaging quality.
As shown in Table \ref{tab:vbench} and Table \ref{tab:vbench_long}, Attn-Forcing shows more pronounced gains in long-horizon generation.
Specifically, on 30s videos, Attn-Forcing achieves the best total and quality scores of 83.85 and 85.26, while achieving the second-best semantic score of 78.18.
On the more challenging 60s generation, Attn-Forcing achieves the best total score of 82.37, with a dynamic degree of 72.14 and the best aesthetic and imaging quality scores, consistently surpassing existing baselines and demonstrating its effectiveness in preserving generation quality over extended durations.
Qualitative comparisons further confirm the advantage of our approach.
As shown in Figure \ref{fig:main_result}, Attn-Forcing produces more dynamic long videos while maintaining competitive temporal consistency and visual quality over extended durations.

\begin{figure*}
    \centering
    \includegraphics[width=1.0\linewidth]{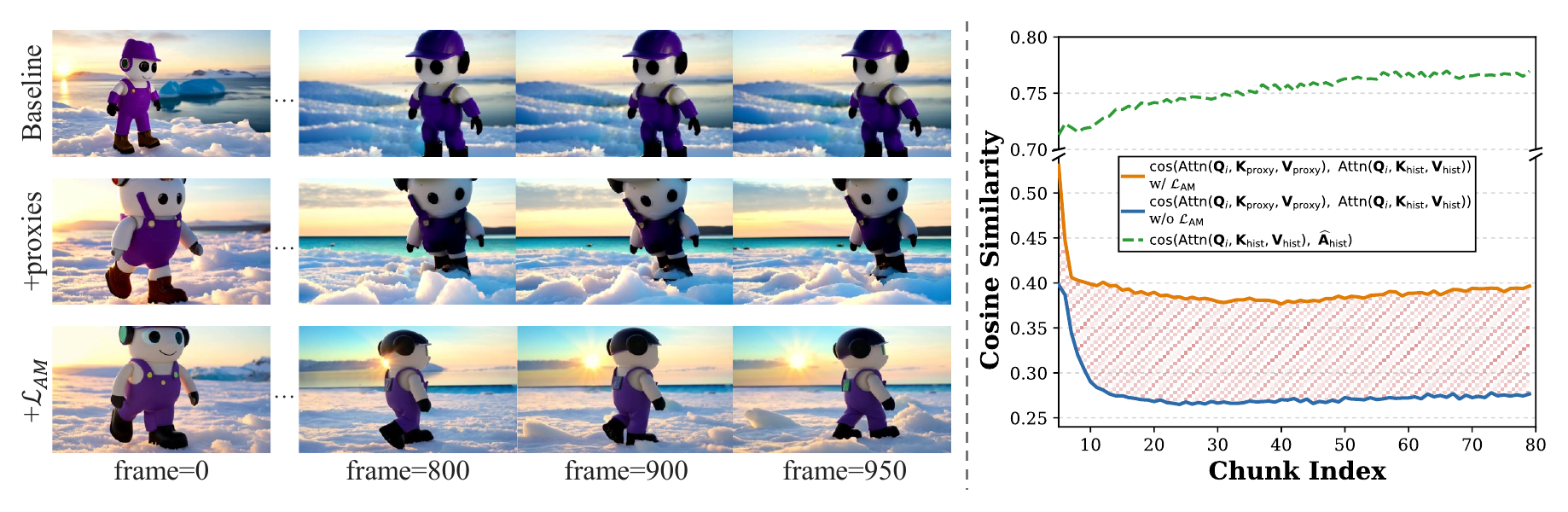}
    \vspace{-8mm}
    \caption{(\textbf{Left}) Qualitative comparison of ablations on our proposed modules. (\textbf{Right}) The orange and blue curves show the cosine similarity between proxy and exact historical attention with and without $\mathcal{L}_{\mathrm{AM}}$, respectively. The green dashed curve shows the similarity between exact historical attention and its approximation $\widehat{\mathbf{A}}_{\mathrm{hist}}$. }
     \vspace{-4mm}
    \label{fig:ablation}
\end{figure*}

\subsection{Ablation Studies}

\begin{wraptable}{r}{0.6\textwidth}
    \centering
    \vspace{-7mm}
    \setlength{\tabcolsep}{4pt}
    \renewcommand{\arraystretch}{1.0}
    \caption{Ablation studies on key components using 60s videos generated from the 128 MovieGen prompts. The best results for the ``Improvement'' module are indicated in \textbf{bold}.}
    \label{tab:ablation}

    \resizebox{\linewidth}{!}{
    \begin{tabular}{@{}lccccc@{}}
        \toprule
        \multirow{2}{*}{Model}
        & \multicolumn{5}{c}{Evaluation Scores on \textbf{60s} $\uparrow$} \\
        \cmidrule(l){2-6}
        & Total & Background & Smooth & Dynamic & Quality \\
        \midrule

        \rowcolor{gray!20}
        \multicolumn{6}{@{}l}{\textit{Improvement}} \\

        Baseline
        & 80.03
        & \textbf{96.25}
        & \textbf{98.57}
        & 46.30
        & 68.70 \\

        + proxies
        & 80.16
        & 96.13
        & 98.35
        & 46.82
        & 69.41 \\

        + $\mathcal{L}_{\mathrm{AM}}$
        & \textbf{82.37}
        & 95.84
        & 98.56
        & \textbf{72.14}
        & \textbf{69.42} \\

        \midrule
        \rowcolor{gray!20}
        \multicolumn{6}{@{}l}{\textit{Impact of $M$}} \\

        $M=128$
        & 82.37
        & 95.84
        & 98.56
        & 72.14
        & 69.42 \\

        $M=64$
        & 79.50
        & 95.89
        & 98.48
        & 47.08
        & 67.52 \\

        $M=32$
        & 79.31
        & 95.69
        & 98.27
        & 47.03
        & 68.05 \\

        \midrule
        \rowcolor{gray!20}
        \multicolumn{6}{@{}l}{\textit{Impact of $L$}} \\

        $L=12$
        & 80.23
        & 95.00
        & 98.20
        & 65.31
        & 67.06 \\

        $L=6$
        & 82.37
        & 95.84
        & 98.56
        & 72.14
        & 69.42 \\

        $L=3$
        & 79.63
        & 95.89
        & 98.46
        & 45.10
        & 69.11 \\

        \bottomrule
    \end{tabular}
    }
    \vspace{-3mm}
\end{wraptable}
We conduct ablation studies on the key components of our framework using 60s videos generated from the 128 MovieGen prompts, as follows:

\noindent{\textbf{Impact of proxies and $\mathcal{L}_{\mathrm{AM}}$.}}
We first apply the standard streaming long tuning as the baseline.
Starting from the baseline, introducing learnable attention proxies improves the image quality score from 68.70 to 69.41, while slightly increasing the total score from 80.03 to 80.16, as shown in Table~\ref{tab:ablation}.
Further incorporating $\mathcal{L}_{\mathrm{AM}}$ substantially improves the total score to 82.37, with a particularly notable gain in dynamic degree from 46.82 to 72.14.
Visual results in Figure \ref{fig:ablation} (Left) also show that our full model maintains high generation quality and strong dynamics throughout one minute of generation.
These qualitative and quantitative comparisons demonstrate the effectiveness of Attn-Forcing.
Moreover, to directly assess the effect of $\mathcal{L}_{\mathrm{AM}}$, we compare the cosine similarity between 
$\operatorname{Attn}\left(\mathbf{Q}_i,\mathbf{K}_{\mathrm{proxy}},\mathbf{V}_{\mathrm{proxy}}\right)$ and $\operatorname{Attn}\left(\mathbf{Q}_i,\mathbf{K}_{\mathrm{hist}},\mathbf{V}_{\mathrm{hist}}\right)$ with and without $\mathcal{L}_{\mathrm{AM}}$ at the last DiT layer, and plot the similarity curves over a 60s generation horizon.
As shown in Figure~\ref{fig:ablation} (Right), incorporating $\mathcal{L}_{\mathrm{AM}}$ (the orange curves) consistently yields higher attention similarity throughout the generation horizon, demonstrating that it effectively improves the approximation of historical attention responses by the learned proxies.

\noindent{\textbf{Impact of the number of random features $M$.}}
The number of random features $M$ controls the approximation capacity of the historical attention statistics.
As shown in Table~\ref{tab:ablation}, reducing $M$ from 128 to 64 or 32 leads to a clear performance drop, with the total score decreasing from 82.37 to 79.50 and 79.31, respectively.
The degradation is particularly significant in dynamic degree, which drops from 72.14 to around 47, while the image quality score also decreases from 69.42 to 67.52 and 68.05.
Meanwhile, background consistency and motion smoothness remain relatively stable.
Therefore, we set $M=128$ in our final model.

\noindent{\textbf{Impact of the number of selected layers $L$.}}
We further study the number of DiT layers $L$ supervised by $\mathcal{L}_{\mathrm{AM}}$.
As shown in Table~\ref{tab:ablation}, $L=6$ achieves the best overall performance, with a total score of 82.37 and a dynamic degree of 72.14.
Accordingly, we use the last 6 DiT layers in our final model.

\subsection{Analysis}

\noindent{\textbf{Full-context attention and $\widehat{\mathbf{A}}_{\mathrm{hist}}$ for generation.}}
We further investigate the effect of directly using
\begin{wraptable}{r}{0.6\textwidth}
    \centering
    \vspace{-5mm}
    \setlength{\tabcolsep}{2pt}
    \renewcommand{\arraystretch}{1.0}
    \caption{Quantitative comparison of LongLive, LongLive with a local attention size of 120, and Attn-Forcing. The best results are indicated in \textbf{bold}.}
    \label{tab:analysis}

    \resizebox{\linewidth}{!}{
    \begin{tabular}{@{}lccccc@{}}
        \toprule
        \multirow{2}{*}{Model}
        & \multicolumn{5}{c}{Evaluation Scores on \textbf{60s} $\uparrow$} \\
        \cmidrule(l){2-6}
        & Total & Background & Smooth & Dynamic & Quality \\
        \midrule
        LongLive
        & 79.26
        & 96.60
        & 98.85
        & 35.55
        & 68.81 \\

        w/ Attn. Size = 120
        & 75.01
        & \textbf{97.08}
        & \textbf{99.37}
        & 11.69
        & 57.27 \\
        
        \midrule
        
        Attn-Forcing
        & \textbf{82.37}
        & 95.84
        & 98.56
        & \textbf{72.14}
        & \textbf{69.42} \\

        \midrule
        \bottomrule
    \end{tabular}
    }
    \vspace{-3mm}
\end{wraptable}
full-context attention or the approximated historical attention response $\widehat{\mathbf{A}}_{\mathrm{hist}}$ during long-horizon generation.
We first extend the local attention size of LongLive from 12 to 120 during 60s video inference (240, corresponding to full-context attention, exceeds the available GPU memory).
As shown in Table~\ref{tab:analysis}, this improves background consistency and motion smoothness from 96.60 and 98.85 to 97.08 and 99.37, respectively.
However, the total score drops from 79.26 to 75.01, with dynamic degree decreasing from 35.55 to 11.69 and image quality from 68.81 to 57.27.
This degradation may stem from LongLive being trained with a much shorter attention window due to GPU memory constraints, limiting its ability to handle longer contexts at inference.

\begin{wrapfigure}{r}{0.6\linewidth}
		\centering
        \vspace{-5mm}
		\includegraphics[width=\linewidth]{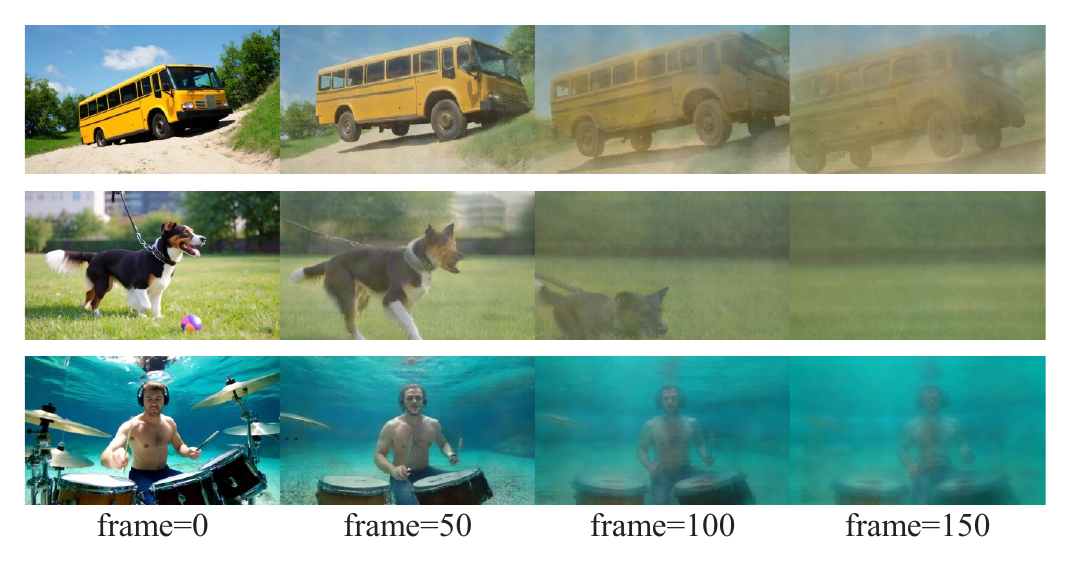}
        \vspace{-8mm}
\caption{Qualitative results of directly using $\widehat{\mathbf{A}}_{\mathrm{hist}}$ for generation. Direct use of the approximation results in increasing blur over time.}
\vspace{-3mm}
\label{fig:approximation}
\end{wrapfigure}
We further replace the proxy attention with $\widehat{\mathbf{A}}_{\mathrm{hist}}$ during inference.
Although $\widehat{\mathbf{A}}_{\mathrm{hist}}$ serves as a good approximation to the exact historical attention response (as illustrated by the green dashed curve in Figure~\ref{fig:ablation} (Right)), directly using this approximation leads to increasing blur as generation proceeds, as shown in Figure~\ref{fig:approximation}.
This is because $\widehat{\mathbf{A}}_{\mathrm{hist}}$ is obtained through random-feature approximation rather than native KV states, introducing accumulation errors into the inference-time attention computation.
In contrast, our attention proxies produce native KV states through the standard DiT forward pass while preserving full-context attention contributions.

\noindent{\textbf{Inference time analysis.}}
Compared with LongLive, Attn-Forcing introduces additional computation for proxy updates and one extra DiT forward pass.
In practice, these operations incur only approximately 13\% additional end-to-end inference time for generating a video on an H20 GPU.

\section{Conclusion}
In this work, we presented Attn-Forcing, a novel framework for preserving full-context attention contributions in streaming video generation under bounded memory and computation.
We achieve this by introducing learnable attention proxies that continuously incorporate evicted visual context and produce native KV states for subsequent generation.
To further improve their approximation of historical attention responses, we proposed an Attention Matching Loss with efficient full-context supervision constructed from online accumulated statistics.
Extensive experiments on both short- and long-horizon video generation demonstrate that Attn-Forcing consistently improves generation quality and achieves state-of-the-art performance over existing baselines, highlighting the effectiveness of explicitly modeling historical attention contributions for high-quality streaming video generation.

\clearpage

\bibliography{iclr2027_conference}
\bibliographystyle{iclr2027_conference}

\clearpage

\appendix
\section{Appendix}
\subsection{Derivation of the Positive Random Feature Approximation}
\label{app:prf}
We provide the derivation of the positive random feature approximation used in
Section~\ref{sec:aml}.
Following~\cite{choromanski2020rethinking}, we consider the softmax kernel
\begin{equation}
\kappa(\mathbf{q},\mathbf{k})
=
\exp\left(
\frac{\mathbf{q}^{\top}\mathbf{k}}{\sqrt{d_h}}
\right).
\end{equation}
We absorb the attention scaling into the
query and key by defining
\begin{equation}
\bar{\mathbf{q}}
=
\frac{\mathbf{q}}{d_h^{1/4}},
\qquad
\bar{\mathbf{k}}
=
\frac{\mathbf{k}}{d_h^{1/4}},
\end{equation}
such that
\begin{equation}
\kappa(\mathbf{q},\mathbf{k})
=
\exp\left(
\bar{\mathbf{q}}^{\top}\bar{\mathbf{k}}
\right).
\end{equation}

Let
$\boldsymbol{\omega}\sim\mathcal{N}(\mathbf{0},\mathbf{I})$
and define the positive random feature
\begin{equation}
\phi_{\boldsymbol{\omega}}(\mathbf{q})
=
\exp\left(
\boldsymbol{\omega}^{\top}\bar{\mathbf{q}}
-
\frac{\|\bar{\mathbf{q}}\|_2^2}{2}
\right),
\qquad
\phi_{\boldsymbol{\omega}}(\mathbf{k})
=
\exp\left(
\boldsymbol{\omega}^{\top}\bar{\mathbf{k}}
-
\frac{\|\bar{\mathbf{k}}\|_2^2}{2}
\right).
\label{eq:app_prf}
\end{equation}
We first rewrite the softmax kernel as
\begin{align}
\kappa(\mathbf{q},\mathbf{k})
&=
\exp\left(
\bar{\mathbf{q}}^{\top}\bar{\mathbf{k}}
\right)
\nonumber\\
&=
\exp\left(
-\frac{\|\bar{\mathbf{q}}\|_2^2}{2}
\right)
\exp\left(
\frac{\|\bar{\mathbf{q}}+\bar{\mathbf{k}}\|_2^2}{2}
\right)
\exp\left(
-\frac{\|\bar{\mathbf{k}}\|_2^2}{2}
\right).
\label{eq:app_softmax_rewrite}
\end{align}
For any $\mathbf{c}\in\mathbb{R}^{d_h}$,
\begin{equation}
(2\pi)^{-d_h/2}
\int
\exp\left(
-\frac{\|\boldsymbol{\omega}-\mathbf{c}\|_2^2}{2}
\right)
d\boldsymbol{\omega}
=
1.
\end{equation}
Setting
$\mathbf{c}=\bar{\mathbf{q}}+\bar{\mathbf{k}}$
gives
\begin{align}
\exp\left(
\frac{\|\bar{\mathbf{q}}+\bar{\mathbf{k}}\|_2^2}{2}
\right)
&=
(2\pi)^{-d_h/2}
\int
\exp\left(
-\frac{\|\boldsymbol{\omega}\|_2^2}{2}
+
\boldsymbol{\omega}^{\top}
(\bar{\mathbf{q}}+\bar{\mathbf{k}})
\right)
d\boldsymbol{\omega}
\nonumber\\
&=
\mathbb{E}_{\boldsymbol{\omega}\sim
\mathcal{N}(\mathbf{0},\mathbf{I})}
\left[
\exp\left(
\boldsymbol{\omega}^{\top}\bar{\mathbf{q}}
\right)
\exp\left(
\boldsymbol{\omega}^{\top}\bar{\mathbf{k}}
\right)
\right].
\label{eq:app_gaussian_identity}
\end{align}
Substituting Eq.~\ref{eq:app_gaussian_identity} into
Eq.~\ref{eq:app_softmax_rewrite}, we obtain
\begin{equation}
\kappa(\mathbf{q},\mathbf{k})
=
\mathbb{E}_{\boldsymbol{\omega}}
\left[
\phi_{\boldsymbol{\omega}}(\mathbf{q})
\phi_{\boldsymbol{\omega}}(\mathbf{k})
\right],
\label{eq:app_prf_identity}
\end{equation}
which is the positive random feature representation used in
Eq.~\ref{eq:kernel_expectation}.

Given $M$ sampled random projections
$\{\boldsymbol{\omega}_m\}_{m=1}^{M}$,
we define the finite-dimensional feature mapping
\begin{equation}
\phi(\mathbf{q})
=
\frac{1}{\sqrt{M}}
\begin{bmatrix}
\phi_{\boldsymbol{\omega}_1}(\mathbf{q}) \\
\vdots \\
\phi_{\boldsymbol{\omega}_M}(\mathbf{q})
\end{bmatrix},
\qquad
\phi(\mathbf{k})
=
\frac{1}{\sqrt{M}}
\begin{bmatrix}
\phi_{\boldsymbol{\omega}_1}(\mathbf{k}) \\
\vdots \\
\phi_{\boldsymbol{\omega}_M}(\mathbf{k})
\end{bmatrix}.
\end{equation}
Therefore,
\begin{equation}
\phi(\mathbf{q})^{\top}\phi(\mathbf{k})
=
\frac{1}{M}
\sum_{m=1}^{M}
\phi_{\boldsymbol{\omega}_m}(\mathbf{q})
\phi_{\boldsymbol{\omega}_m}(\mathbf{k})
\approx
\kappa(\mathbf{q},\mathbf{k}),
\end{equation}
which gives Eq.~\ref{eq:kernel_approx}.
In implementation, following~\cite{choromanski2020rethinking}, we use Gaussian orthogonal random projections, which maintain the Gaussian marginal distribution while reducing the variance of the finite-sample kernel estimator.

\subsection{Training and Inference Procedure}
\label{app:training_inference}
\begin{algorithm}[t]
\caption{Training Procedure}
\label{alg:attn_training}
\begin{algorithmic}[1]
\REQUIRE Autoregressive chunks $\{\mathbf{x}_i\}_{i=1}^{N}$,
proxy latents $\mathbf{P}$, selected last $L$ DiT layers
\STATE Initialize $\mathbf{P}$ and
$\{\mathbf{S}_{\mathrm{hist}}^{l},\mathbf{z}_{\mathrm{hist}}^{l}\}$

\FOR{$i=1,\ldots,N$}

    \IF{historical statistics are available}

    \STATE
    $(\mathbf{K}_{\mathrm{proxy}},\mathbf{V}_{\mathrm{proxy}})
    \xleftarrow{\,G_{\theta}\ \mathrm{forward}\,}
    \mathbf{P}$

        \STATE
        $\hat{\mathbf{x}}_i,\{\mathbf{Q}_i^l\}
        \leftarrow
        G_{\theta}
        (\mathbf{x}_{i,t};
        \mathbf{KV}_{\mathrm{sink}},
        \mathbf{KV}_{\mathrm{proxy}},
        \mathbf{KV}_{\mathrm{local}})$

        \FOR{$l$ in the last $L$ DiT layers}
            \STATE
            $\widehat{\mathbf{A}}_{\mathrm{hist}}^{l}
            \leftarrow
            \left[
            \operatorname{Diag}
            \big(
            \Phi(\mathbf{Q}_i^l)
            \mathbf{z}_{\mathrm{hist}}^l
            \big)
            \right]^{-1}
            \Phi(\mathbf{Q}_i^l)
            \mathbf{S}_{\mathrm{hist}}^l$

            \STATE
            $\mathbf{A}_{\mathrm{proxy}}^{l}
            \leftarrow
            \operatorname{Attn}
            (\mathbf{Q}_i^l,
            \mathbf{K}_{\mathrm{proxy}}^l,
            \mathbf{V}_{\mathrm{proxy}}^l)$
        \ENDFOR

        \STATE Compute $\mathcal{L}_{\mathrm{AM}}$
        using Eq.~\ref{eq:attention_matching_loss}

    \ELSE
        \STATE
        $\hat{\mathbf{x}}_i
        \leftarrow
        G_{\theta}
        (\mathbf{x}_{i,t};
        \mathbf{KV}_{\mathrm{sink}},
        \mathbf{KV}_{\mathrm{local}})$
        \STATE $\mathcal{L}_{\mathrm{AM}}\leftarrow 0$
    \ENDIF

    \IF{visual latents are evicted}
        \STATE
        $\mathbf{P}
        \leftarrow
        \mathcal{U}
        (\mathbf{P},\mathbf{x}_{i}^{\mathrm{evict}})$
        \hfill (Eq.~\ref{eq:proxy_update})

        \FOR{$l$ in the last $L$ DiT layers}
            \STATE
            $\mathbf{S}_{\mathrm{hist}}^l
            \leftarrow
            \mathbf{S}_{\mathrm{hist}}^l
            +
            \Phi(\mathbf{K}_{\mathrm{evict}}^l)^{\top}
            \mathbf{V}_{\mathrm{evict}}^l$

            \STATE
            $\mathbf{z}_{\mathrm{hist}}^l
            \leftarrow
            \mathbf{z}_{\mathrm{hist}}^l
            +
            \Phi(\mathbf{K}_{\mathrm{evict}}^l)^{\top}\mathbf{1}$
        \ENDFOR
    \ENDIF

\ENDFOR

\STATE
$\mathcal{L}
\leftarrow
\mathcal{L}_{\mathrm{DMD}}
+
\lambda_{\mathrm{AM}}\mathcal{L}_{\mathrm{AM}}$
\hfill (Eq.~\ref{eq:overall_loss})

\STATE Update model parameters using $\mathcal{L}$

\end{algorithmic}
\end{algorithm}

\begin{algorithm}[t]
\caption{Inference Procedure}
\label{alg:attn_inference}
\begin{algorithmic}[1]
\REQUIRE Autoregressive chunks $\{\mathbf{x}_i\}_{i=1}^{N}$,
proxy latents $\mathbf{P}$

\FOR{$i=1,\ldots,N$}

    \IF{visual latents have been evicted}
        \STATE
        $(\mathbf{K}_{\mathrm{proxy}},
        \mathbf{V}_{\mathrm{proxy}})
        \xleftarrow{\,G_{\theta}\ \mathrm{forward}\,}
        \mathbf{P}$

        \STATE
        $\hat{\mathbf{x}}_i
        \leftarrow
        G_{\theta}
        (\mathbf{x}_{i,t};
        \mathbf{KV}_{\mathrm{sink}},
        \mathbf{KV}_{\mathrm{proxy}},
        \mathbf{KV}_{\mathrm{local}})$
    \ELSE
        \STATE
        $\hat{\mathbf{x}}_i
        \leftarrow
        G_{\theta}
        (\mathbf{x}_{i,t};
        \mathbf{KV}_{\mathrm{sink}},
        \mathbf{KV}_{\mathrm{local}})$
    \ENDIF

    \IF{visual latents are evicted}
        \STATE
        $\mathbf{P}
        \leftarrow
        \mathcal{U}
        (\mathbf{P},\mathbf{x}_{i}^{\mathrm{evict}})$
        \hfill (Eq.~\ref{eq:proxy_update})
    \ENDIF

\ENDFOR
\end{algorithmic}
\end{algorithm}
We summarize the training and inference procedures of Attn-Forcing in
Algorithms~\ref{alg:attn_training} and~\ref{alg:attn_inference}, respectively.
During streaming long tuning, we first perform standard DMD training until visual latents begin to be evicted from the local attention window.
Once eviction occurs, the evicted states are incorporated into our proxy training process.
Specifically, the corresponding visual latents are used to update the proxy latents through the Proxy Update Module, while their cached layer-wise KV states are used to update the historical attention statistics.
Starting from subsequent chunks, the updated proxy latents are propagated through the standard DiT forward pass to obtain their layer-wise proxy KV states, which are then inserted between the sink and local KV states for attention computation.
Meanwhile, for the last $L=6$ DiT layers, the accumulated statistics are used to construct the historical attention supervision for $\mathcal{L}_{\mathrm{AM}}$.
The resulting $\mathcal{L}_{\mathrm{AM}}$ is combined with the DMD objective according to Eq.~\ref{eq:overall_loss}.
We optimize the proxy-related parameters with $\mathcal{L}_{\mathrm{AM}}$ only during generator updates.

At inference time, Attn-Forcing follows the autoregressive streaming generation process.
Before any visual latents are evicted from the local attention window, each chunk is generated using the sink and local KV states.
Once eviction occurs, the corresponding visual latents are used to update the proxy latents through the Proxy Update Module.
Starting from the next chunk, the updated proxy latents are propagated through the DiT to obtain the proxy KV states, which are inserted between the sink and local KV states for attention computation.
As generation proceeds, newly evicted visual latents continuously update the proxy latents for subsequent chunks.
The historical attention statistics and $\mathcal{L}_{\mathrm{AM}}$ are used only during training and are not required at inference.

\begin{figure*}
    \centering
    \includegraphics[width=1.0\linewidth]{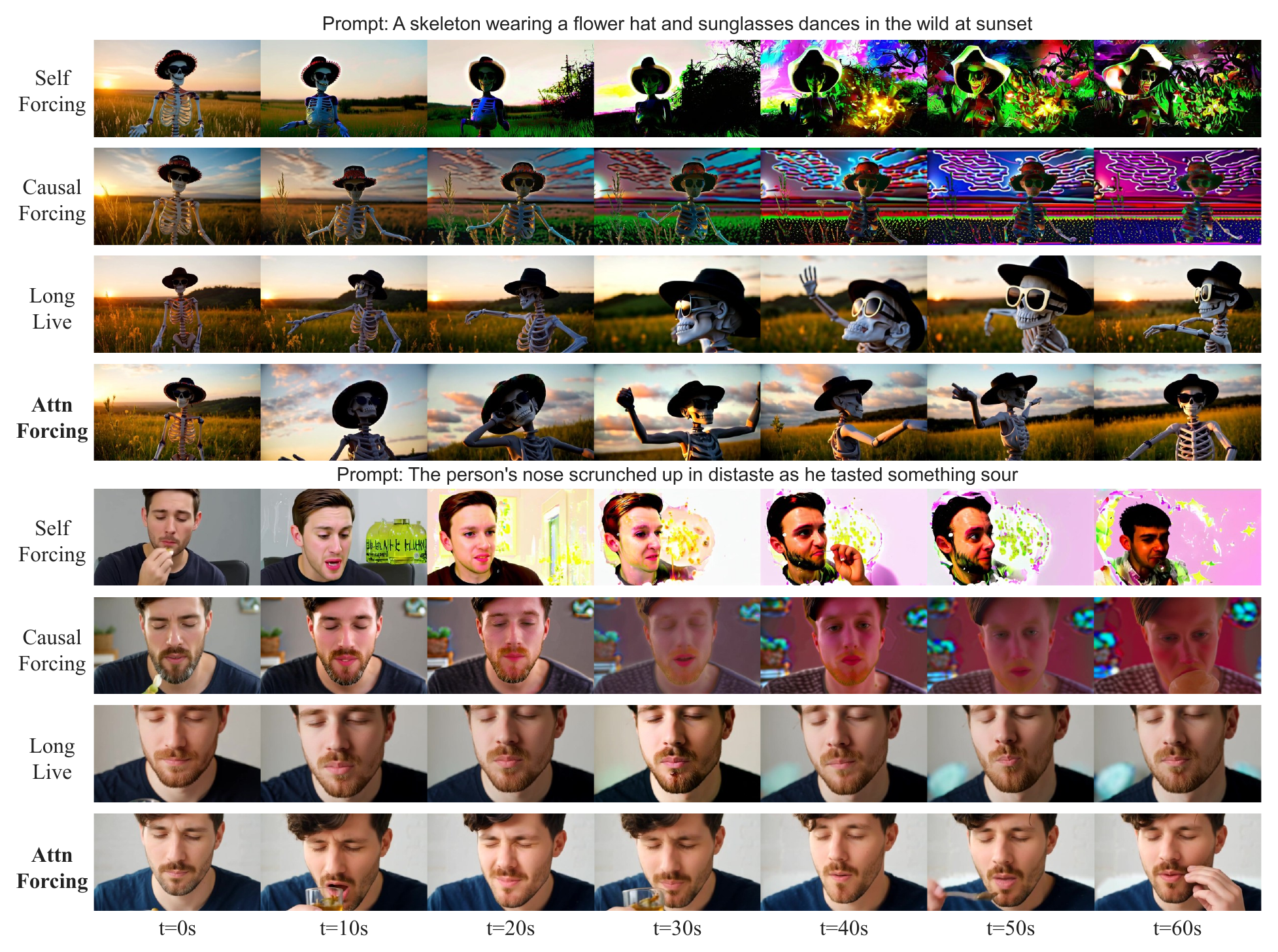}
    \vspace{-8mm}
    \caption{Additional qualitative comparison on 60s videos generated using MovieGen prompts. Attn-Forcing achieves stronger generation quality and motion dynamics. }
     \vspace{-4mm}
    \label{fig:add_result}
\end{figure*}

\subsection{Implementation Details}
\label{app:proxy_update}

We provide additional details of the Proxy Update Module $\mathcal{U}$ in Eq.~\ref{eq:proxy_update}.
For simplicity, we omit the module-specific superscripts of the learnable projections.
Given the current proxy latents $\mathbf{P}$ and newly evicted visual hidden
states $\mathbf{x}^{\mathrm{evict}}$, we first summarize the evicted context
through cross-attention:
\begin{equation}
\mathbf{H}
=
\operatorname{Attn}
\left(
\mathbf{W}_{Q}(\mathbf{P}),
\mathbf{W}_{K}(\mathbf{x}^{\mathrm{evict}}),
\mathbf{W}_{V}(\mathbf{x}^{\mathrm{evict}})
\right).
\end{equation}
The resulting context is then further refined and fused with the current proxy
latents through an additional cross-attention block followed by an FFN:
\begin{align}
\mathbf{C}
&=
\mathbf{P}
+
\operatorname{Attn}
\left(
\mathbf{W}_{Q}(\mathbf{P}),
\mathbf{W}_{K}(\mathbf{H}),
\mathbf{W}_{V}(\mathbf{H})
\right),
\nonumber\\
\mathbf{C}
&\leftarrow
\mathbf{C}
+
\operatorname{FFN}
\left(
\operatorname{LN}(\mathbf{C})
\right).
\end{align}

The refined context is further transformed into a candidate update:
\begin{equation}
\boldsymbol{\Delta}
=
\operatorname{RMSNorm}
\left(
\operatorname{MLP}(\mathbf{C})
\right),
\end{equation}
where $\operatorname{MLP}(\cdot)$ consists of two linear layers with a GELU
activation.

We then compute an element-wise gate
\begin{equation}
g
=
\sigma
\left(
\mathbf{W}
\left[
\operatorname{LN}(\mathbf{P});
\operatorname{LN}(\mathbf{C})
\right]
+
\mathbf{b}
\right),
\end{equation}
and obtain the updated proxy latents through the gated update module
$\mathcal{G}$:
\begin{equation}
\mathcal{G}
\left(
\mathbf{P},
\mathbf{H}
\right)
=
g\odot\mathbf{P}
+
(1-g)\odot\boldsymbol{\Delta}.
\label{eq:app_proxy_update}
\end{equation}
For temporal positional encoding, the proxies are assigned the same RoPE positions as the sink frames.
For gradient flow, the current queries in $\mathcal{L}_{\mathrm{AM}}$ and the evicted visual hidden states used for proxy updates are treated as stop-gradient inputs.

\subsection{Additional Qualitative Results}
In this subsection, we present additional visual comparison on 60s videos in Figure \ref{fig:add_result}.

\subsection{Limitations}
Although Attn-Forcing preserves historical attention contributions under bounded memory and computation, several limitations remain.
First, Attn-Forcing maintains a fixed number of attention proxies throughout generation.
While this design keeps the additional memory and computational overhead bounded, it also imposes a fixed capacity for representing the accumulated history.
As the generation horizon grows substantially longer or the visual content becomes increasingly diverse, a limited number of proxies may become insufficient to preserve all relevant historical information.

Second, the Attention Matching Loss relies on a random-feature approximation of the historical attention response rather than exact full-context attention.
Although this approximation enables efficient long-context supervision using constant-size online statistics, it inevitably introduces approximation error into the supervision signal.

\end{document}